\documentclass[sigconf,natbib=true]{acmart}

\pdfoutput=1
\makeatletter
\def\mdseries@tt{m}
\makeatother 
\usepackage[draft=true,newfloat]{minted} 
\usepackage{graphicx}
\usepackage{hyperref}           

\renewcommand\footnotetextcopyrightpermission[1]{} 
\makeatletter
\renewcommand\@formatdoi[1]{\ignorespaces}
\makeatother

\usepackage[utf8]{inputenc}
\usepackage{float}
\usepackage{setspace}
\usepackage{placeins}
\usepackage{listings}
\usepackage{caption}
\newenvironment{code}{\captionsetup{type=listing}}{}
\SetupFloatingEnvironment{listing}{name=Listings}
\usepackage{pgfplots}
\usepgfplotslibrary{groupplots}
\usepackage{xcolor,pifont}
\definecolor{amber}{rgb}{1.0, 0.75, 0.0}
\definecolor{awesome}{rgb}{1.0, 0.13, 0.32}
\definecolor{ao(english)}{rgb}{0.0, 0.5, 0.0}

\newcommand{\specialcell}[2][c]{%
  \begin{tabular}[#1]{@{}c@{}}#2\end{tabular}}

\acmConference{Graph For All Million Dollar Challenge 2022}

\title{TigerLily: Finding drug interactions in silico with the \textit{Graph}}

\author{Benedek Rozemberczki}
\affiliation{
  \institution{}
  \country{United Kingdom}
}

\begin{document}

\maketitle

\section{Introduction}
Adverse drug-drug interactions cause every year thousands of fatalities in the United States \cite{craigle2007medwatch} and lead to the hospitalization of many more; one can only imagine these numbers for the whole world. This extreme number of unnecessary death and burden on the health care system could be avoided if better drug-drug interaction indications would be available to drug discovery researchers, clinicians, and patients. Moreover, adverse drug interaction events affect the elderly population and those excluded from private healthcare systems and self-medicating disproportionately \cite{hines2011potentially, berreni2015adverse}. In a time when the population of many countries is aging \cite{bjorkman2002drug} and self-medication is becoming increasingly widespread \cite{bennadi2013self} there is an urgent need for reliable drug-drug interaction indications that are available to all. 

\begin{figure}[h!]
\centering 
\includegraphics[scale=0.12]{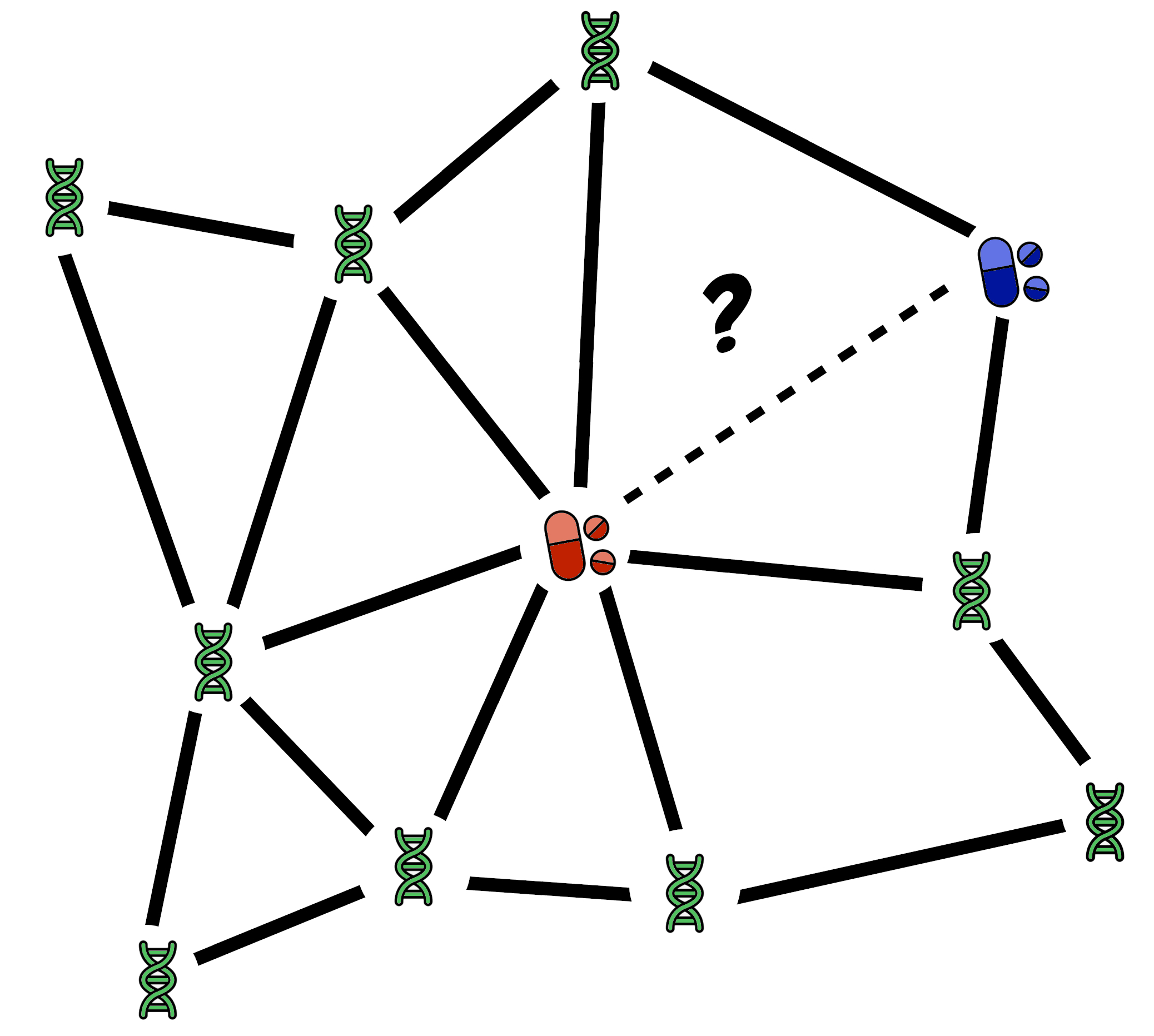}
\caption{TigerLily solves the in silico drug interaction prediction problem \cite{rozemberczki2021unified} using a heterogeneous biological graph \cite{biosnapnets}. Given information about systematic interactions of drugs (red and blue pills) and gene targets (green double helix segments) our goal is to find unexpected potentially dangerous links between drugs.  }\label{fig:eyecandy}
\end{figure}

However, collecting data about potential drug-drug interaction indications is an extremely challenging task due to multiple reasons: (a) the number of potential interactions increases quadratically with the number of drugs; (b) experimental validation is time-consuming and costly concerning equipment; (c) experimental results can be noisy and inconsistent across laboratories and patients; (d) the observed symptoms might be unrelated to the parallel administration of drugs. The reasons listed above make computational approaches to the drug-drug interaction predicting problem, particularly appealing in practical settings \cite{rozemberczki2021unified,rozemberczki2022chemicalx}.

The proposed framework TigerLily offers a computational solution to the drug-drug interaction prediction problem based on systems biology and graphs \cite{kitano2002systems}. A cartoon-like graphical summary of the main idea behind TigerLily is depicted in Figure \ref{fig:eyecandy}. Based on a biological network where edges are interactions, drugs and genes are nodes our goal is to predict interactions. We assume that based on the \textit{biological network} based neighbourhood context of drugs one can predict the drug-drug interactions. TigerLily learns to embed the drug nodes in a feature space using personalized PageRank scores computed with TigerGraph. Using the drug embedding features drug pair features are defined which serve as input to supervised drug interaction classifiers which learn from known interactions and can predict novel adverse events.

\subsection{Statement of Significance}
Releasing TigerLily is significant for multiple reasons. We aim to briefly summarize these reasons and how the release of TigerLily is related to the main goals of the \textit{Graph For All Million Dollar Challenge}. 

\subsubsection{Impactfulness} Drug-drug interactions affect everyone who has to take multiple medications in parallel, a tool that can indications of them can benefit drug discovery researchers, clinicians, patients, and drug safety regulators. A widely applicable and accessible tool that can indicate adverse interactions can reduce the number of fatalities and hospitalization rates. 
\subsubsection{Innovativeness} TigerLily uses a novel node embedding technique that is based on pruned approximate Personalized PageRank scores. This node embedding is innovative, because: (i) the technique was not described explicitly in the literature previously; (ii) exploits the existing TigerGraph ecosystem (iii) does not require the computation and network transfer of the whole dense personalized PageRank matrix.

\subsubsection{Ambitiousness} The experimental evaluation of TigerLily uses a heterogeneous biological graph with two node types: drugs and genes. This graph has nearly a million edges, more than 1000 drugs, and 20,000 genes -- which nearly covers the whole human genome. The personalized PageRank computation in TigerGraph Cloud exploits the sparsity and graph heterogeneity offered by TigerGraph to reduce the memory and computation requirements of TigerLily.
\subsubsection{Applicability} Throughout the development of TigerLily we followed a pragmatic software engineering approach: the code base is covered by unit and integration tests, documented, continuous integration runs on the repository, and the library is pip installable and we provided tutorials for the users. The solution can be repurposed to solve other tasks such as the drug synergy prediction problem with little effort.

\subsection{Summary of Contributions} Introducing TigerLily makes several significant contributions to the field of graph machine learning-based drug-drug interaction prediction. The main contributions can be summarized as:

\begin{itemize}
    \item We release TigerLily an open-source TigerGraph-based system designed to predict drug-drug interactions in silico using heterogeneous biological graphs.
    \item We evaluate the performance of TigerLily using real-world biological and chemical data that we integrated from DrugBank \cite{wishart2006drugbank, wishart2008drugbank} and BioSNAP \cite{biosnapnets}. 
    \item We discuss those who would benefit from TigerLily, potential limitations to the approach, obstacles for the adoption, and potential future directions for development.
\end{itemize}
The remainder of this project report is structured as follows. In Section \ref{sec:related_work} we overview the relevant literature about drug-drug interactions, proximity preserving node embeddings, and biological knowledge graphs. Formal definitions of graph mining and the underlying mathematical model behind TigerLily are discussed in Section \ref{sec:preliminaries}. We focus on the practical design of the framework in Section \ref{sec:design} with code snippets with a real-world running example. The system is evaluated in Section \ref{sec:experiments} with real-world data. The limitations of Tigerlily are discussed in Section \ref{sec:limitations} and the report concludes in Section \ref{sec:conclusions} with potential future directions for research and development. The library is available under \url{https://github.com/benedekrozemberczki/tigerlily}.

\section{Related work}\label{sec:related_work}
Our high-level overview of the related work primarily aims to position TigerLily in the existing literature about drug-drug interaction prediction, heterogeneous biological graphs, and node embedding algorithms.
\subsection{Drug-Drug Interaction Prediction}
The drug-drug interaction prediction problem is part of the wider drug pair scoring task \cite{rozemberczki2021unified, rozemberczki2022chemicalx}. In this one has to assign labels to a pair of drugs that describes the behaviour of the drugs in a biological context. This context can be drug-drug interaction \cite{ryu2018deep}, polypharmacology side effects \cite{zitnik2018modeling} or the synergy of the drugs when administered together \cite{rozemberczki2021moomin}. According to \cite{rozemberczki2021unified} machine learning-based solutions to this challenge can be categorized into three main groups: (a) molecular features based models \cite{ryu2018deep} (b) network biology-based embedding models \cite{zitnik2018modeling}  (c) hierarchical models which use a mixture of molecular features and systems biology \cite{rozemberczki2021moomin}. Our work is closest to the network biology-based solutions as it exploits a high-level biological entity graph to solve the problem by creating upstream drug embeddings and a downstream classifier.

\subsection{Heterogeneous Biological Graphs}
A heterogeneous biological graph consists of biological entities (node types) and heterogeneous interactions between these entities (edge types). Publicly available graphs \cite{bonner2021review} are differentiated by each other based on the types of nodes and edges present in the graph, various node type hierarchies \cite{bonner2021review}, the size of the biological graph \cite{edwards2021explainable} and the use case that was driving the creation of the graph \cite{geleta2021biological}. These heterogeneous biological graphs can be used by drug discovery researchers in the pharmaceutical industry \cite{bonner2021review} as an input for off- and on target drug repurposing \cite{gaudelet2021utilizing}, gene target identification \cite{gogleva2022knowledge}, and compound interaction prediction \cite{rozemberczki2021moomin} systems.

\subsection{Node Embeddings}
Node embeddings are unsupervised machine learning models which map the nodes of a graph into an Euclidean space \cite{qiu2018network} where various notions of graph-based proximity between nodes (e.g. neighbourhood overlap \cite{grover2016node2vec}, personalized PageRank \cite{pagerank}, adjacency \cite{rozemberczki2018fast}) are preserved. By doing this for each node a feature vector is assigned which can be used to solve various downstream machine learning task such as node classification \cite{grover2016node2vec}, link prediction \cite{rozemberczki2018fast, grover2016node2vec} or node clustering \cite{rozemberczki2018fast}. The drug-drug interaction problem that is our interest can be formulated as a link prediction task on a heterogeneous biological graph, because of this we are going to take a customized node embedding-based approach.

\section{Preliminaries}\label{sec:preliminaries}
In this section, we focus on the mathematical model that powers the predictions made by TigerLily and the biological graph dataset that we created to test the performance of the machine learning system. \footnote{The embedding model described in Subsection \ref{subsec:theory} is involved technically, it gives reasoning to the reader why the proposed solution interfaces well with the architecture of TigerGraph. The reader can jump ahead and read the rest of the report if such details seem less relevant to the scope of the challenge.}
\subsection{The Embedding Model}\label{subsec:theory}
TigerLily relies on a custom node embedding model which exploits the advantageous functionalities of TigerGraph. Our goal is to give a concise and prompt description of this upstream machine learning model and how the features of this model are used by the downstream drug interaction predictor.
\subsubsection{Graph Theory Basics.} Drugs and genes are noted by the sets $\mathcal{D}$ and  $\mathcal{P}$. The union of these two sets $\mathcal{V}=\mathcal{D} \cup \mathcal{P}$ defines the node set (biological entities), $\mathcal{E}$ is the set of edges between the entities and $G=(\mathcal{V},\mathcal{E})$ is the heterogeneous biological graph. We postulate that the edge set does not contain any drug-drug edges hence $(d,d')\notin \mathcal{E}, \forall d, d' \in \mathcal{D}$. We use $n=|\mathcal{V}|$ and $m=|\mathcal{D}|$ to denote the cardinality of the biological entity and drug sets. Finally, $\tilde{\textbf{A}}$ is the $n \times n$ normalized adjacency matrix of the graph $G$.
\subsubsection{Upstream Drug Embeddings.} A drug indicator matrix $\textbf{S}\in \mathbb{R}^{n\times m}_{(0,1)}$ is a binary matrix where each row corresponds to a biological entity and columns correspond to drugs. Non zero entries of this matrix correspond to indicators of the drugs in the graph, meaning that $\forall v \in \mathcal{V}, d \in \mathcal{D}$ it only holds that $\textbf{S}_{v,d} = 1$ if $v=d$. The approximate personalized PageRank scores \cite{pagerank,predict2019klicpera,he2020lightgcn} of drugs are defined by Equation \eqref{eq:PPR}.
\begin{align}
\textbf{X}&=\sum \limits_{r=0}^{t-1}\alpha \cdot (1-\alpha)^r \widetilde{\textbf{A}}^r \textbf{S}+(1-\alpha)^t \widetilde{\textbf{A}}^t \textbf{S}\label{eq:PPR}
\end{align}
Here $r$ is a running index, $t$ is the number of approximation iterations,  $0\leq\alpha\leq 1$ is the return probability, and $\textbf{X}\in\mathbb{R}^{n \times m}_{0+}$ is the matrix of approximate personalized PageRank scores for the drug nodes in the graph. An entry in this matrix is large when a source biological entity (row) is close to a drug (column) based on the approximate personalized PageRank score.

Let us define $\widetilde{\textbf{X}}\in\mathbb{R}^{m \times n}_{0+}$ the pruned approximate personalized PageRank matrix between drugs and biological entities by Equation \eqref{eq:prune}. The pruning operation takes a matrix as an input and returns a sparse matrix wherein each row the top $k$ largest values are kept, everything else is zeroed out. An entry in this matrix is large when a source drug (row) is close to a biological entity (column) based on the pruned approximate personalized PageRank score. It must be emphasized that this matrix only takes negligible $\mathcal{O}(mk)$ space compared to the $\mathcal{O}(n^2)$ space required by personalized PageRank for all biological entities.
\begin{align}
\widetilde{\textbf{X}}&=\textbf{PRUNE}(\textbf{X}^\top, k)\label{eq:prune}
\end{align}

This sparse matrix can be easily computed by using the personalized PageRank query of TigerGraph on the biological graph. Finally, we can learn the drug embeddings from this matrix by solving the non-negative matrix factorization problem \cite{nmf_1,nmf_2} defined by Equation \eqref{eq:optim}.
\begin{align}
\min \left\| \widetilde{\textbf{X}} - \textbf{H}\textbf{W} \right\|_F \text{ subject to } \textbf{H} \in \mathbb{R}^{m\times d}_{0+}, \textbf{W}\in \mathbb{R}^{d\times n}_{0+}  \label{eq:optim}
\end{align}
In Equation \eqref{eq:optim} $d<<m$ is the number of embedding dimensions, the non negative matrix $\textbf{H}$ is the drug embedding and $\textbf{W}$ is the biological entity embedding. Each row of $\textbf{H}$ is a drug and columns can be interpreted as hidden features of the embedding; we are going to use the features of this matrix as drug features to define the drug pair features and to train the interaction predictor.
\subsubsection{Downstream Drug Pair Classifier} Given the drug set $\mathcal{D}$ and drug embedding matrix $\textbf{H}$ the feature vector describing the drug pair $d, d'\in \mathcal{D}$ is  defined as $\textbf{H}_{(d,d')} = g(\textbf{H}_d,\textbf{H}_{d'})$ where the function $ g(\cdot)$ is a so called operator function \cite{grover2016node2vec}; an example operator function can be the concatenation of the two drug vectors or the Hadamard product of the vectors. A downstream classifier takes such drug pair feature vector for a pair of drugs $d,d'\in\mathcal{D}$ as an input and outputs the probability that there is an interaction between them.

\subsection{The Integrated Drug Interaction Dataset}
Our experiments use a manually integrated biological network from BioSNAP \cite{biosnapnets} and a DrugBank DDI \cite{ryu2018deep} based drug pair dataset.

\subsubsection{The Biological Graph}
We took all of the available non-cell-specific gene-gene and drug-gene interaction networks \cite{zitnik2018modeling, szklarczyk2016stitch} from BioSNAP and integrated them into a single heterogeneous biological graph. Genes have been mapped to the Entrez identifier system \cite{maglott2005entrez} and drugs have been mapped to DrugBank identifiers \cite{wishart2006drugbank} in the data cleaning process. The result is an undirected heterogeneous graph with two node types; 1,106 drug nodes and 20,754 gene nodes, 38,393 drug-gene target interactions, and 778,290 gene-gene regulatory interactions. 

\subsubsection{The Drug-Drug Interactions.}
The target drug-drug interactions were taken from the DrugBank DDI dataset \cite{ryu2018deep}; we filtered for those drug pairs where both drugs are present in the biological network we created. After this filtration we have 187,850 labeled drug pairs; 106,362 of these pairs have an adverse interaction (positive label) and 81,488 of these have no known interaction (negative label). This dataset was made available via the GitHub repository of TigerLily and the library also has a built-in data loader class to access this integrated dataset.

\section{The TigerLily Framework Design}\label{sec:design}

Our discussion about the Tigerlily design focuses on two things: (a) a real-world drug-drug interaction prediction use case that showcases the API step-by-step and interfacing with other machine learning libraries; (b) the software engineering principles that ensure that the TigerGraph based solution is maintainable and robust.
\subsection{A Real World Use Case}

\begin{figure*}[h!]
\centering 
\includegraphics[scale=0.7]{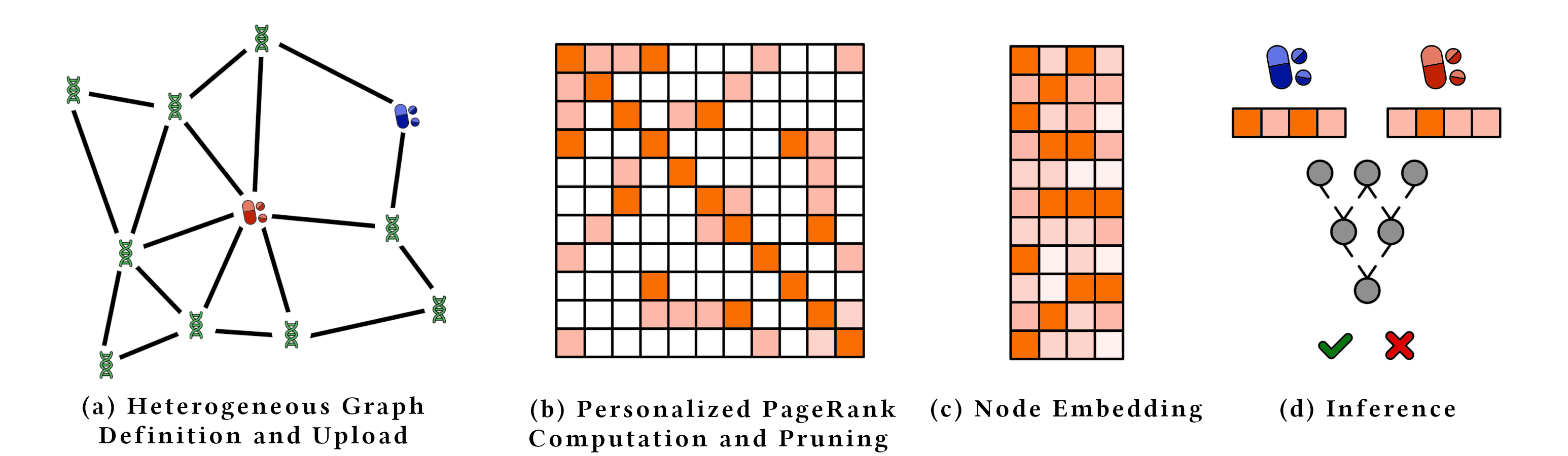}
\caption{TigerLily provides a workflow for biological graph-based drug interaction prediction that consists of multiple main steps. (a) Using TigerGraph a heterogeneous systems biology graph is defined with drug and gene nodes. This graph does not contain the drug-drug interactions. (b) The Personalized PageRank vector for each drug node is computed using TigerGraph and this vector is pruned to contain the top-k most similar nodes based on proximity.  (c) Based on the Personalized PageRank matrix we learn drug embeddings which serve as \textit{Systems Biology} based features of drugs. (d) Using the drug features we define drug pair features and train a classifier to predict the interactions of pair combinations. }\label{fig:main_eyecandy}
\end{figure*}

In this section, we solve a real-world drug-drug interaction problem that uses data from DrugBank and BioSNAP. The details of these datasets and the performance TigerLily on this problem are discussed in Section \ref{sec:experiments} in great detail. A high-level step-by-step overview of the TigerLily based solution to this problem is described in Figure \ref{fig:main_eyecandy} and our demonstration will follow this workflow. 

\subsubsection{Heterogeneous Graph Definition and Upload.} Our goal is to create a heterogeneous biological graph and store it as a drug-gene interaction network with TigerGraph. The way we achieve this is described by the code snippet in Listings \ref{code:a}.

\begin{code}
\begin{minted}[linenos,fontsize=\small,xleftmargin=0.5cm,numbersep=3pt,frame=lines]{python}
from tigerlily.dataset import ExampleDataset
from tigerlily.embedding import EmbeddingMachine
from tigerlily.operator import hadamard_operator
from tigerlily.pagerank import PersonalizedPageRankMachine

dataset = ExampleDataset()

edges = dataset.read_edges()
target = dataset.read_target()

machine = PersonalizedPageRankMachine(host="host",
                                      graphname="graph",
                                      secret="secret",
                                      password="password")
                           
machine.connect()
machine.install_query()

machine.upload_graph(new_graph=True, edges=edges)
\end{minted}
\captionof{listing}{Loading the example drug-drug interaction dataset, creating a TigerLily \texttt{PersonalizedPageRankMachine} instance and populating the Graph with a heterogeneous biological graph.}\label{code:a}
\end{code}

We start by importing classes and functions from TigerLily that we will use later (lines 1-4). We create a \texttt{ExampleDataset} instance, load the edges and the target drug pairs with the respective class methods. Both of these parts of the dataset are returned as \texttt{pandas dataframe} objects by the class methods (lines 6-9). We must note, that the \texttt{edges} dataframe must have columns named \texttt{node\_1}, \texttt{node\_2}, \texttt{drug\_1} and \texttt{drug\_2} columns which respectively contain the node identifiers and the node types. The edge dataframe must not contain edges where both of the nodes have a drug node type as our goal is to predict the existence of drug-drug interactions. In a similar fashion the \texttt{target} dataframe must have the columns \texttt{drug\_1}, \texttt{drug\_2} and \texttt{label} which contain the drug identifiers and the indicator for the existence or non-existence of an interaction.

We create a \texttt{PersonalizedPageRankMachine} instance (lines 11-14) that is parametrized by the hostname and graph names with the appropriate credentials. This step requires that there is a running and existing TigerGraph Cloud instance with a graph that has the appropriate Graph schema - in our case, it means drug and gene nodes with interaction edges. The instance is connected, the personalized PageRank query is installed and the edges of the graph are uploaded (lines 16-19). By setting the \texttt{new\_graph} flag to be true we make sure that the graph is empty before the upload starts. After the graph is populated it is time to compute some approximate personalized PageRank scores.

\subsubsection{Personalized PageRank Computation and Pruning.} Our focus is on the drug nodes and we want to query those to get an understanding of which are those biological entities that are in close proximity to the drugs. We do this with the piece of code described in Listings \ref{code:b}. First, we query the \texttt{PersonalizedPageRankMachine} to get a list of the drug nodes (line 1). Based on this list using the \texttt{get\_personalized\_pagerank} class method we query those nodes for each drug that have a large approximate personalized PageRank score. For each drug we return the top-$k$ highest scoring entry (line 3); this ensures that the returned dataset is small even when the number of biological entities is large compared to the number of drugs. The returned \texttt{pagerank\_scores} data frame has the \texttt{node\_1}, \texttt{node\_2} and \texttt{score} columns; it describes a sparse matrix where rows are drugs, columns are biological entities (including drugs) and the values are pruned personalized PageRank scores. Let us move on to the learning drug embeddings from this matrix and creating drug pair features from the learned embeddings.

\begin{code}
\begin{minted}[linenos,fontsize=\small,xleftmargin=0.5cm,numbersep=3pt,frame=lines]{python}
drugs = machine.connection.getVertices("drug")

pagerank_scores = machine.get_personalized_pagerank(drugs)
\end{minted}
\captionof{listing}{Querying the previously created  TigerLily \texttt{PersonalizedPageRankMachine} instance for the list of drug nodes in the Graph and computing the personalized PageRank of nodes in the proximity of drugs.}\label{code:b}
\end{code}

\subsubsection{Drug Node Embedding.} Using the previously computed personalized PageRank matrix we learn node embeddings for each drug with the Python script in Listings \ref{code:c}. We create an \texttt{EmbeddingMachine} instance and learn drug node embeddings (lines 1-5); the returned \texttt{embedding} is a a \texttt{pandas} dataframe where the firs column named \texttt{node\_id} contains the drug identifiers and the remaining columns contain the embedding dimensions. Using the \texttt{target} and the drug pair feature computation \texttt{hadamard\_operator} the \texttt{create\_features} class method of the \texttt{EmbeddingMachine} instance allows the creation of drug pair features (lines 7-8). Using these drug pair features we are ready to solve the drug-drug interaction task!

\begin{code}
\begin{minted}[linenos,fontsize=\small,xleftmargin=0.5cm,numbersep=3pt,frame=lines]{python}
embedding_machine = EmbeddingMachine(seed=42,
                                     dimensions=32,
                                     max_iter=100)

embedding = embedding_machine.fit(pagerank_scores)

features = embedding_machine.create_features(target,
                                   hadamard_operator)
\end{minted}
\captionof{listing}{Creating a \texttt{EmbeddingMachine} instance, learning drug embeddings from the personalized PageRank scores and creating drug pair features with the \texttt{hadamard\_operator} for the drug pairs.}\label{code:c}
\end{code}

\subsubsection{Drug Pair Interaction Prediction and Inference.} Our final step to learning a drug pair classifier and producing predictions for the drug pairs is summarized with code in Listings \ref{code:d}. This step is a pretty generic supervised machine learning workflow based on LightGBM \cite{ke2017lightgbm} and scikit-learn \cite{pedregosa2011scikit}.

We start by importing the gradient boosted classifier, the AUROC evaluation metric, and the function for creating train-test splits (lines 1-3). We split the drug pair features and the labels with the \texttt{train\_test\_split} function into training and testing parts (line 5). We create a \texttt{LightGBMClassifier} instance, learn the model from the training set drug pairs, score on the test set, compute the AUROC score and print the score by taking the first few digits (lines 7-17). This snippet demonstrated that TigerLily interfaces with existing machine learning libraries smoothly.

\begin{code}
\begin{minted}[linenos,fontsize=\small,xleftmargin=0.5cm,numbersep=3pt,frame=lines]{python}
from lightgbm import LGBMClassifier
from sklearn.metrics import roc_auc_score
from sklearn.model_selection import train_test_split

X_train, X_test, y_train, y_test = train_test_split(features,
                                                    target)

model = LGBMClassifier(learning_rate=0.01,
                       n_estimators=100)

model.fit(X_train, y_train["label"])

y_hat = model.predict_proba(X_test)

auroc_score_value = roc_auc_score(y_test["label"],
                          y_hat[:,1])

print(f'AUROC score: {auroc_score_value :.4f}')
\end{minted}
\captionof{listing}{Splitting the target and the TigerLily generated drug pair features to train and evaluate a gradient boosting based drug pair scoring model.}\label{code:d}
\end{code}

\subsection{Maintaining and Supporting TigerLily}
As we have seen in the previous section TigerLily was engineered with an end-user-friendly API in mind and this design is supported by continuous integration, documentation, tutorials, package indexing, and unit- and integration tests.

\subsubsection{Documentation and Example Notebook.} The complete codebase of  TigerLily is documented with docstrings and type annotations. Using these and restructured text files we automatically release new documentation that reflects the current state of the TigerLily Github repository. This documentation is available under \url{https://tigerlily.readthedocs.io/en/latest/} and it covers the API reference and includes a tutorial for the new potential users. The same tutorial is available in the form of a Jupyter Notebook in the repository that explains line by line a typical Tigerlily-based graph analytics workflow. 

\subsubsection{Inclusion in the Python Package Index.} The submission 0.1.0 release of Tigerlily is publicly available on the Python Package Index. This means that different versions of the library (including the submission release) can be accessed via the \url{https://tigerlily.readthedocs.io/} webpage and that it can be installed via the command line using the \texttt{pip install tigerlily} command. This allows the end-users to install TigerLily in the Python environment that they use efficiently and also the same users can install different versions of the library based on their needs.

\subsubsection{Test Suite and Code Coverage Reports.} The continuous integration of Tigerlily with Github Actions allows the automated testing of the codebase. The Tigerlily namespaces are covered by unit and integration tests which ensure that software components behave as expected. Each automated test suite run generates a coverage report hosted on Codecov. These reports are publicly available under \url{https://app.codecov.io/gh/benedekrozemberczki/tigerlily} and allow inspecting the coverage rate of the TigerLily namespaces.

\section{Experimental Evaluation}\label{sec:experiments}

The main goal of this section is to demonstrate that TigerLily can solve a real-world problem. Using the biological network and drug-drug interaction data discussed in Section \ref{sec:preliminaries} we will test the predictive performance of classifiers that use TigerLily-based drug embedding features.

\subsection{Predictive Performance}\label{subsec:pred}

The predictive efficacy of TigerLily is a primary driver of the potential impact that the solution can have in the real world. Because of this we investigate this and compare the performance under various drug pair feature computation operators with multiple binary classification metrics.
\subsubsection{Experimental Design} We compute approximate personalized PageRank scores with the \texttt{PersonalizedPageRankMachine} for the drugs and their closest 50 neighbors, from 25 PageRank approximation iterations with a return probability of 0.7. Drug embeddings are learned with the \texttt{EmbeddingMachine} in 32 dimensions by doing 100 iterations. From the drug embeddings, drug pair features are computed with the operators listed in Table \ref{tab:performance}. We use 80\% of the pairs to train a gradient boosted tree machine (LightGBM implementation \cite{ke2017lightgbm}) and compute AUROC, AUPR, F$_1$ scores on the remaining 20\% of pairs. The average performance computed from 10 random seeded experimental runs is in Table \ref{tab:performance} with standard errors around the mean performance.

\begin{table}[h!]
\caption{The mean predictive performance (standard deviations below) of a TigerLily based gradient boosted machine computed from 10 seeded train-test splits. Rows represent drug-drug feature computation routines described by  \cite{grover2016node2vec}.}\label{tab:performance}
\begin{tabular}{ccccc}
\hline
\specialcell{\textbf{Binary}\\\textbf{Operator} }      & \specialcell{\textbf{Definition} \\ $\textbf{of component H}_{(d,d')}^i $}& \textbf{AUROC} & \textbf{AUPR} & $\textbf{F}_1$ \\ \hline
\textbf{Absolute}              &  $|\textbf{H}^i_d-\textbf{H}_{d'}^i|$          &      $\underset{\pm .002}{.953}$ &   $\underset{\pm .002}{.961}$   &  $\underset{\pm .003}{.874}$  \\
\textbf{Squared}            &        $(\textbf{H}^i_d-\textbf{H}_{d'}^i)^2$     &     $\underset{\pm .001}{.951}$  &   $\underset{\pm .002}{.963}$   &  $\underset{\pm .002}{.870}$  \\
\textbf{Difference}      &    $\textbf{H}^i_d-\textbf{H}_{d'}^i$        &     $\underset{\pm .003}{.948}$  & $\underset{\pm .002}{.961}$     & $\underset{\pm .004}{.860}$   \\
\textbf{Hadamard}       &    $\textbf{H}^i_d\cdot\textbf{H}_{d'}^i$       &    $\underset{\pm .003}{.951}$   &    $\underset{\pm .002}{.955}$  & $\underset{\pm .003}{.871}$   \\ \hline
\end{tabular}
\end{table}

\subsubsection{Experimental Results} Most importantly the results in Table \ref{tab:performance} are a strong signal that TigerLily-based embedding features can be used to predict the drug-drug interactions. We can also observe that there is no clearly superior operator and that there is a negligible difference in performance between the various feature computation operators across the performance metrics. A further error analysis could show which drugs participate in drug pairs that are being consistently misclassified by TigerLily. 

\subsection{Training Data Ratio}
Getting ground truth about known drug-drug interactions is a costly process. Because of this, data-efficient solutions which require a limited amount of labeled drug pairs to solve the drug-drug interaction problem are extremely valuable. We are going to investigate the predictive performance of TigerLily under various training data ratio regimes with a range of downstream classifiers.

\subsubsection{Experimental Design} 
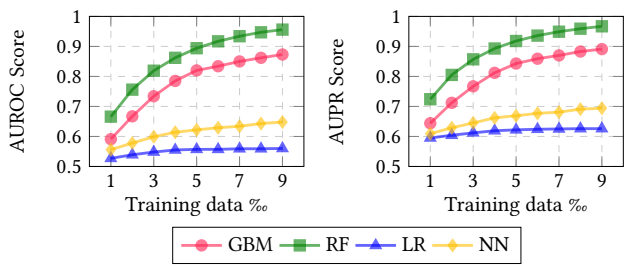
\begin{figure}[h!]
\centering
\scalebox{0.9}{
\begin{tikzpicture}
\begin{groupplot}[	grid=major,
	grid style={dashed, gray!40},group style={
                      group name=myplot,
                      group size= 2 by 1, horizontal sep=1.55cm,vertical sep=1.2cm},height=3.8cm,width=4.75cm, title style={at={(0.5,0.9)},anchor=south},every axis x label/.style={at={(axis description cs:0.5,-0.15)},anchor=north},]
\nextgroupplot[
	ylabel=AUROC Score,
	xlabel=Training data \textperthousand,
	xtick={1,3,5,7,9},
	xmin=0,
	xmax=10,
	ymin=0.5,
	ymax=1.0,
	ytick={0.5,0.6,0.7,0.8,0.9,1.0},
 	legend columns=4,
	legend style={at={(1.20,-0.65)},anchor=south},
y label style={at={(0.05,0.5)}},
    legend entries={GBM,RF,LR,NN},
]
\addplot [very thick, awesome,mark=*,opacity=0.6]coordinates {
( 1 , 0.591 )
( 2 , 0.667 )
( 3 , 0.734 )
( 4 , 0.785 )
( 5 , 0.820 )
( 6 , 0.834 )
( 7 , 0.850 )
( 8 , 0.862 )
( 9 , 0.873 )
};
\addplot [very thick, ao(english),mark=square*,opacity=0.6]coordinates {
( 1 , 0.666 )
( 2 , 0.756 )
( 3 , 0.819 )
( 4 , 0.862 )
( 5 , 0.894 )
( 6 , 0.917 )
( 7 , 0.934 )
( 8 , 0.947 )
( 9 , 0.956 )
};
\addplot [very thick, blue, ,mark=triangle*,opacity=0.6]coordinates {
( 1 , 0.527 )
( 2 , 0.539 )
( 3 , 0.548 )
( 4 , 0.555 )
( 5 , 0.557 )
( 6 , 0.557 )
( 7 , 0.559 )
( 8 , 0.559 )
( 9 , 0.560 )
};
\addplot [very thick, amber,mark=diamond*,opacity=0.6]coordinates {
( 1 , 0.556 )
( 2 , 0.578 )
( 3 , 0.599 )
( 4 , 0.614 )
( 5 , 0.622 )
( 6 , 0.629 )
( 7 , 0.634 )
( 8 , 0.643 )
( 9 , 0.648 )
};

\nextgroupplot[
	xtick={1,3,5,7,9},
    y label style={at={(0.05,0.5)}},
	ylabel=AUPR Score,
	xlabel=Training data \textperthousand,
	xtick={1,3,5,7,9},
	xmin=0,
	xmax=10,
	ymin=0.5,
	ymax=1.0,
	ytick={0.5,0.6,0.7,0.8,0.9,1.0},
	]
\addplot [very thick, awesome,mark=*,opacity=0.6]coordinates {
( 1 , 0.644 )
( 2 , 0.712 )
( 3 , 0.767 )
( 4 , 0.812 )
( 5 , 0.843 )
( 6 , 0.859 )
( 7 , 0.870 )
( 8 , 0.883 )
( 9 , 0.891)
};
\addplot [very thick, ao(english),mark=square*,opacity=0.6]coordinates {
( 1 , 0.724 )
( 2 , 0.805 )
( 3 , 0.857 )
( 4 , 0.893 )
( 5 , 0.918 )
( 6 , 0.936 )
( 7 , 0.949 )
( 8 , 0.959 )
( 9 , 0.967 )
};
\addplot [very thick, blue, ,mark=triangle*,opacity=0.6]coordinates {
( 1 , 0.595 )
( 2 , 0.604 )
( 3 , 0.612 )
( 4 , 0.619 )
( 5 , 0.622 )
( 6 , 0.624 )
( 7 , 0.625 )
( 8 , 0.626 )
( 9 , 0.626 )
};
\addplot [very thick, amber,mark=diamond*,opacity=0.6]coordinates {
( 1 , 0.609 )
( 2 , 0.629 )
( 3 , 0.645 )
( 4 , 0.662 )
( 5 , 0.669 )
( 6 , 0.677 )
( 7 , 0.681 )
( 8 , 0.691 )
( 9 , 0.694 )
};

\end{groupplot}
\end{tikzpicture}}
\caption{The mean drug interaction prediction performance of TigerLily based classifiers conditioned on the ratio of training data (in permille) calculated from 10 train-test splits. }\label{fig:data_ratio}
\end{figure}

We train LightGBM \cite{ke2017lightgbm} based gradient boosting and scikit-learn \cite{pedregosa2011scikit} based random forest, logistic regression, and neural networks to predict the interactions using drug pair features computed with the Hadamard operator. The personalized PageRank calculation and embedding hyperparameter settings were taken from Subsection \ref{subsec:pred}. We modulate the amount of training data and plot the test set average predictive performance computed from 10 random seeded experimental runs on the subplots of Figure \ref{fig:data_ratio} as a function of the training data amount.

\subsubsection{Experimental Results} The line charts of Figure \ref{fig:data_ratio} clearly demonstrate that the non-parametric methods (gradient boosting and random forest) are extremely data-efficient compared to the parametric ones. Given less than 1\% of data, these methods achieve similar results to the number in Table \ref{tab:performance}. This showcases that the two-stage machine learning system of TigerLily is able to solve the drug interaction problem with high predictive efficacy even when the amount of training data is extremely limited.

\subsection{Embedding Dimension Sensitivity}

TigerLily is a highly modular framework with the upstream drug embeddings and downstream drug pair classifier-based design. However, the number of dimensions used for the drug embeddings is a highly important hyperparameter of the upstream model; in a certain sense, it is a key hyperparameter to tune and optimize.

\subsubsection{Experimental Design} 
\begin{figure}[h!]

	\centering
	\begin{tikzpicture}[scale=0.45,transform shape]
	\tikzset{font={\fontsize{16pt}{12}\selectfont}}
	\begin{groupplot}[group style={group size=2 by 2,
		horizontal sep=60pt, vertical sep=60pt,ylabels at=edge left},
	width=0.54\textwidth,
	height=0.3375\textwidth,
	ymin=0.48,
	ymax=1.02,
	legend columns=3,
every tick label/.append style={font=\bf},
    y tick label style={
        /pgf/number format/.cd,
            fixed,
            fixed zerofill,
            precision=0,
        /tikz/.cd
    },
 enlarge x limits=true,
	grid=major,
	grid style={dashed, gray!40},
	scaled ticks=false,
	inner axis line style={-stealth}]

 \nextgroupplot[
   xlabel=$\log_2$ Embedding dimensions,
    ybar=0pt,
      every tick/.style={
        black,
        semithick,
      },
    bar width=9pt,
    enlargelimits=0.17,
    ylabel={AUROC Score},
    legend style={at={(0.5,-0.15)},
      anchor=north,legend columns=-1},
    symbolic x coords={2,3,4,5,6, 7},
    xtick={2,3,4,5,6, 7},
yticklabels={0.5,0.6,0.7,0.8,0.9,1.0},
ytick={0.5,0.6,0.7,0.8,0.9,1.0},
    ]

\addplot [fill=awesome!55,draw=awesome,error bars/.cd,y dir=both,y explicit]  coordinates {
(2,0.842)
(3,0.888) 
(4,0.920) 
(5,0.948) 
(6,0.960) 
(7,0.968) 
};
\addplot [fill=blue!55,draw=blue,error bars/.cd,y dir=both,y explicit]  coordinates {
(2,0.509) 
(3,0.511) 
(4,0.537) 
(5,0.558) 
(6,0.605)
(7,0.723) 

};
\addplot [fill=amber!55,draw=amber,error bars/.cd,y dir=both,y explicit] coordinates {

(2,0.516) 
(3,0.576) 
(4,0.618) 
(5,0.779) 
(6,0.882) 
(7,0.891) 
};

 \nextgroupplot[
   xlabel=$\log_2$ Embedding dimensions,
    ybar=0pt,
      every tick/.style={
        black,
        semithick,
      },
    bar width=9pt,
    enlargelimits=0.17,
    legend columns=4,
    legend image post style={solid},
    legend style={at={(0.5,-0.25)},nodes={scale=1.5, transform shape}, 
      anchor=north,legend columns=-1},
    ylabel={AUPR Score},
yticklabels={0.5,0.6,0.7,0.8,0.9,1.0},
ytick={0.5,0.6,0.7,0.8,0.9,1.0},
    symbolic x coords={2,3,4,5,6, 7},
    xtick={2,3,4,5,6, 7},
    	legend style = { column sep = 10pt, legend columns = 1, legend to name = grouplegend, font=\small}  ]

\addplot [fill=awesome!55,draw=awesome,error bars/.cd,,y dir=both,y explicit]  coordinates {
(2,0.871) 
(3,0.908) 
(4,0.934) 
(5,0.953) 
(6,0.970)
(7,0.976)
};\addlegendentry{GBM}

\addplot [fill=blue!55,draw=blue,error bars/.cd,y dir=both,y explicit]   coordinates {
(2,0.570) 
(3,0.573)
(4,0.601) 
(5,0.618) 
(6,0.681) 
(7,0.708)};\addlegendentry{LR}

\addplot [fill=amber!55,draw=amber,error bars/.cd,y dir=both,y explicit] coordinates {
(2,0.575)
(3,0.626)
(4,0.672)
(5,0.803) 
(6,0.882) 
(7, 0.907)
};\addlegendentry{NN}
	\end{groupplot}

	\node at ($(group c1r1) + (5.0cm,-4.2cm)$) {\ref{grouplegend}}; 
	\end{tikzpicture}
	
	\caption{The mean drug interaction prediction performance of TigerLily feature based classifiers conditioned on the number of embedding features calculated from 10 train-test splits. }\label{fig:barchart}
\end{figure}
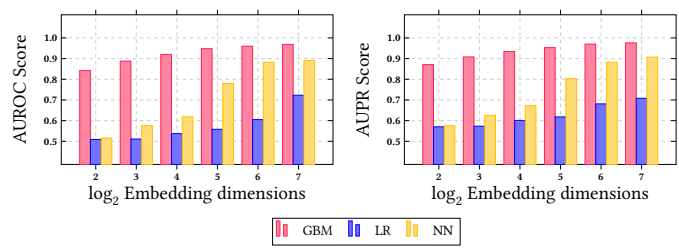

Using the personalized PageRank, upstream embedding and downstream model settings from the previous section we train gradient boosted tree, logistic regression, and neural network classifiers while the number of embedding dimensions is modulated in $\left\{2^2,\dots,2^7\right\}$. We compute average AUROC and AUPR scores from 10 random seeded experimental runs and plotted the average predictive performance as a function of the embedding dimensions in Figure \ref{fig:barchart}.

\subsubsection{Experimental Results} Our results in Figure \ref{fig:barchart} demonstrate that increasing the embedding dimensions beyond $2^5$ the default is beneficial for predicting novel drug-drug interactions. We can also observe that the non-parametric gradient boosted trees gain less with the increased number of embedding dimensions compared to the parametric models.

\section{Target Users and Limitations}\label{sec:limitations}

\subsection{Potential Target Users}
When TigerLily was designed we had specific user stories in mind about how it could be deployed. All of these stories showcase the potential of TigerLily to affect the life of millions of patients around the world. 
\subsubsection{Early Drug Discovery Researchers.} Using TigerLily in the early discovery phase could allow drug discovery researchers to flag potentially adverse drug interactions early on. Indications could be followed up by experiments later in the clinical phase of the drug discovery process. This could reduce the attrition rate of drugs being developed as drugs with a lot of adverse interactions would not proceed to the late phases of the development process.

\subsubsection{Clinical Practitioners.} Using TigerLily could give clinical practitioners early warnings about potential interactions of drugs that are prescribed to patients. Based on these warnings the doctors can make informed decisions about the drug combinations. For example, the patients could be asked to look for specific polypharmacy side effects and monitored more closely when there is a risk of adverse drug interactions.

\subsubsection{Pharmaceutical Industry Regulators.} During the drug discovery process the potential drug-drug interactions are not a primary target for the researchers. Because of this potential drug safety concerns can be flagged during the drug approval. A TigerLily-like system that can create indications for potential interactions could serve the pharmaceutical industry regulators who could get early warnings about potential safety issues and rare adverse events.

\subsubsection{Self-medicating Patients.} A large number of adverse drug interaction events happen when patients decide to use multiple drugs simultaneously on their own \cite{asseray2013frequency}. By extending TigerLily with a convenient and friendly user interface patients could make better self-medication decisions after consulting TigerLily. This could be extremely beneficial for those communities that are critically neglected and under-served by the healthcare system.

\subsection{Limitations and Obstacles for Deployment}
Our system TigerLily is a proof of concept for TigerGraph based in silico drug-drug interaction prediction. This means that the project and the presented approach have some limitations which we would like to highlight shortly.

\subsubsection{Systems Biology Data of New Drugs is Lacking.}
The node embedding functionality of TigerLily implicitly assumes that the drugs of interest are connected to the gene-gene network so drugs can be contextualized by their location in the biological graph. This particular issue is a manifestation of the transductive node representation learning setting \cite{hamilton2017inductive} of our embedding approach. However, this is not true for newly developed drugs, where the nature of the exact interactions with other genes is not understood beyond a primary gene target. This means that predicting interactions for compounds that are not known for long can be a challenging task and potentially the performance of TigerLily could be affected. This challenge could be overcome by using inductive graph neural networks \cite{hamilton2017inductive, kipf2017semi,  predict2019klicpera, bojchevski2020scaling,graph2018velickovic,rozemberczki2021pdn} which would require extrinsic drug and gene features.
\subsubsection{Limited Number of Biological Entity Types.}
The graph which we integrated to demonstrate the predictive performance of our solution has two types of biological entities: drugs and gene targets. Existing heterogeneous biological graphs \cite{geleta2021biological,walsh2020biokg,himmelstein2017systematic} used in the pharmaceutical industry have a larger variety of node types such as cell lines, protein variants, biological processes, and pathways. It is possible that the inclusion of more node and edge types would enrich the graph and result in better quality drug embeddings and interaction predictions.

\subsubsection{Skewed Drug Interaction Data.} Drug interaction and synergy databases are known to be skewed \cite{zitnik2018modeling, rozemberczki2021unified, ryu2018deep}. Practically this means that combinations that involve specific commonly used drugs are tested, but combinations of rarely used drugs are not present. Given that the primary goal of in silico drug interaction predictions is to give indications for combinations of rarely used pairs this can be problematic.

\subsubsection{Homogeneous Proximity Preserving Node Embedding.} Our approach creates homogeneous proximity preserving node embeddings based on the approximate Personalized PageRank scores; this is a considerable limitation given our graph. A number of current approaches are able to incorporate information about the heterogeneity of the graph with respect to node types and edges \cite{balazevic2019tucker, nickel2011three, trouillon2017knowledge}. Moreover, certain embedding approaches are able to describe the structural roles of nodes \cite{rossi2020proximity} which could be important when it comes to the systems biology of compounds and genes.

\subsubsection{Poor Interpretability of Predictions.} TigerLily is a framework that consists of an upstream and downstream module; by the end of the upstream phase, the graph information is distilled into feature vectors. Downstream models trained on these features are not interpretable and because of this, we cannot answer questions about why the classifier flagged a drug pair to be a potential adverse one. However, given extrinsic drug and gene features and a graph neural network trained on these modern explanation techniques could allow the creation of such post-hoc explanations of the drug-drug interaction predictions 
\cite{ying2019gnnexplainer,yuan2021explainability, pope2019explainability,schlichtkrull2021interpreting}.

\subsubsection{Hesitance of Potential Users.} Our proposed drug interaction prediction framework TigerLily serves as an intelligent system that can accelerate the work of clinicians, early discovery researchers, and drug safety experts. However, the black-box nature of this system might be something that cannot be overlooked by the end-users. Moreover, the system might have a potential bias that could affect specific groups of people. One such group could be a set of people who share genetic markers which make them more prone to experience a certain type of adverse events due to drug combinations. Issues like this could make the potential users of TigerLily use the system.

\section{Future Directions and Conclusions}\label{sec:conclusions}

\subsection{New Research Directions}
The open-source codebase, modularity, and extensible nature of TigerLily open up opportunities for future research and engineering solutions. We will highlight potential avenues for the further development of the framework which we see to be high impact and low effort.
\subsubsection{User Interface Development} TigerLily is a Python library that requires that the users are familiar with the language and feel comfortable with reading the documentation. Allowing users to run simple queries on the TigerLily-based predicted scores could open up opportunities for a large number of people. Because of this, the development of a user interface where drug pairs associated with adverse events can be queried seems to be a promising and high-impact future research direction.
\subsubsection{Large Biological Graphs} Our demonstration uses a graph that covers most of the human genome but ignores node types such as pathways or cell lines. We postulate that modeling the underlying biology systems biology better with data and encoding biology better can only be achieved via this. Hence, the integration of large publicly available biological graphs in TigerLily seems to be an extremely promising direction for future research. 
\subsubsection{Drug Synergy Predictions} The focus of our discussion about TigerLily is on adverse drug-drug interactions, but the system is sufficiently flexible to allow for solving other tasks. One interesting task defined on drug pairs is synergy prediction a central question of computational oncology research \cite{rozemberczki2021moomin,rozemberczki2022chemicalx}.

\subsection{Concluding Remarks and Summary}
In this project report we discussed TigerLily a TigerGraph-powered open-source machine learning system for in silico drug interaction prediction. We discussed existing artificial intelligence-based approaches to solving this interesting problem and how biological graphs can be used to tackle this challenge. We gave an overview of how we integrated data from public resources to solve this problem and formalized a mathematical model that can exploit biological graphs to solve the task; TigerLily is a conceptualization of this mathematical model. We demonstrated the main features of TigerLily with Python code examples and highlighted the software engineering principles that make the tool robust, reliable, and accessible. By doing extensive experiments we had shown that TigerLily can help to predict drug interactions in the real world. We reviewed potential users of the system, and its limitations and emphasized the most important future research directions.

\bibliographystyle{ACM-Reference-Format}
\bibliography{main}

\end{document}